 \documentclass[pmlr,twocolumn,10pt]{jmlr} 

\usepackage{booktabs}
\usepackage{hyperref}
\usepackage{siunitx}

\usepackage{multirow}
\usepackage{caption}

\newcommand{\equal}[1]{{\hypersetup{linkcolor=black}\thanks{#1}}}

\theorembodyfont{\upshape}
\theoremheaderfont{\scshape}
\theorempostheader{:}
\theoremsep{\newline}

\jmlrvolume{LEAVE UNSET}
\jmlryear{2023}
\jmlrsubmitted{LEAVE UNSET}
\jmlrpublished{LEAVE UNSET}
\jmlrworkshop{Machine Learning for Health (ML4H) 2023} 

 \title[Deep Learning for Parkinson's Disease Eye Tracking Data]{Deep Learning for Time Series Classification of Parkinson's Disease Eye Tracking Data}

\usepackage{comment}

\author{\Name{Gonzalo Uribarri}\equal{These authors contributed equally}
       \Email{uribarri@kth.se}\\ 
       \addr Section of Computational Brain Science, EECS, KTH, Stockholm\\
       DigitalFutures, KTH, Stockholm\\
       Science for Life Laboratory, KTH, Stockholm
      \AND
      \Name{Simon {Ekman von Huth}}\footnotemark[1]
       \Email{simonevh@kth.se}\\ 
       \addr Section of Computational Brain Science, EECS, KTH, Stockholm
       \AND
       \Name{Josefine Waldthaler}
       \Email{josefine.waldthaler@ki.se}\\ 
       \addr Section of Neurology, Department of Clinical Neuroscience, CMM, Karolinska Institute, Stockholm\\ 
       University Hospital Marburg, Department for Neurology, Marburg
       \AND
       \Name{Per Svenningsson}
       \Email{per.svenningsson@ki.se}\\ 
       \addr Section of Neurology, Department of Clinical Neuroscience, CMM, Karolinska Institute, Stockholm
       \AND
       \Name{Erik Fransén}
       \Email{erikf@kth.se}\\ 
       \addr Section of Computational Brain Science, EECS, KTH, Stockholm,\\
       DigitalFutures, KTH, Stockholm\\
      Science for Life Laboratory, KTH, Stockholm}

\begin{document}

\maketitle
\begin{abstract}
Eye-tracking is an accessible and non-invasive technology that provides information about a subject's motor and cognitive abilities. As such, it has proven to be a valuable resource in the study of neurodegenerative diseases such as Parkinson's disease. Saccade experiments, in particular, have proven useful in the diagnosis and staging of Parkinson's disease. However, to date, no single eye-movement biomarker has been found to conclusively differentiate patients from healthy controls. In the present work, we investigate the use of state-of-the-art deep learning algorithms to perform Parkinson's disease classification using eye-tracking data from saccade experiments. In contrast to previous work, instead of using hand-crafted features from the saccades, we use raw $\sim1.5\,s$ long fixation intervals recorded during the preparatory phase before each trial. Using these short time series as input we implement two different classification models, InceptionTime and ROCKET. We find that the models are able to learn the classification task and generalize to unseen subjects. InceptionTime achieves $78\%$ accuracy, while ROCKET achieves $88\%$ accuracy. We also employ a novel method for pruning the ROCKET model to improve interpretability and generalizability, achieving an accuracy of $96\%$. Our results suggest that fixation data has low inter-subject variability and potentially carries useful information about brain cognitive and motor conditions, making it suitable for use with machine learning in the discovery of disease-relevant biomarkers.
\end{abstract}


\section{Introduction}
\subsection{Motivation}
Parkinson's disease (PD) is a neurodegenerative disorder that affects millions of people worldwide. Beyond the classical symptoms of rigidity, tremor and bradykinesia, one of the most common symptoms of PD is visual impairment, which is partly attributed to impaired oculomotor control [\cite{armstrong_visual_2017}]. One of the most extensively studied experimental protocols for examining eye movements is the prosaccade and antisaccade tasks [\cite{waldthaler_cluster_2023}]. Eye movement during these tasks has proven to be a useful diagnostic tool in differentiating neurological conditions with similar symptoms but different pathophysiological substrates, as different patterns of saccadic impairment reflect pathology in corresponding brain regions [\cite{anderson_eye_2013}]. For example, saccade analysis has been shown to increase accuracy in distinguishing PD from atypical Parkinsonian syndromes [\cite{termsarasab_diagnostic_2015}]. 

However, despite considerable effort, no single eye-movement biomarker has been identified that can conclusively indicate PD [\cite{waldthaler_cluster_2023}], limiting the usefulness of eye-tracking data to diagnose the condition. This challenge calls for more advanced analytical techniques, such as machine learning, to aid in the diagnosis and classification of PD based on eye movement data [\cite{przybyszewski_machine_2023}].

\subsection{Related Work}

One of the largest challenges when applying machine learning to the medical domain is the difficulty of accessing large standardized datasets with a sufficient number of patients, especially when the measurements are not part of regular clinical testing [\cite{kelly_key_2019, rajpurkar_ai_2022}]. This problem is further emphasized in the case of neurodegenerative disease, where the inter-patient variability of the data can be large, and we typically do not have access to data from the same individual at early stages of the disease, and especially not before disease onset. A major challenge when trying to apply machine learning for diagnosis is therefore to obtain good generalization to unseen subjects. 

Previous studies have investigated the use of machine learning algorithms to classify and stage PD based on saccade experiment data [\cite{przybyszewski_machine_2023}]. For instance, in [\cite{przybyszewski_multimodal_2016}], authors employed data from reflexive saccades in 10 PD patients to train a decomposition tree algorithm and classify the stage of the disease. They achieved an accuracy of $79\%$, and similar results were found on antisaccade data in [\cite{sledzianowski_measurements_2019}]. Stuart et al. in [\cite{stuart_pro-saccades_2019}] used linear mixed-effects models to show that features from saccade data are a predictor of cognitive decline in PD patients. The authors of [\cite{brien_classification_2023}] used various point and functional features from a pro-/antisaccade experiment, including error rate, to train a voting classifier composed of 3 sub-classifiers: a support vector machine, a logistic regression model, and a random forest. Their classifier achieved $82\%$ of accuracy in discriminating between PD and healthy controls (HC). In [\cite{waldthaler_cluster_2023}], the authors performed an agglomerative hierarchical cluster analysis on saccade experimental data and showed that there are at least two opposing patterns of saccade changes associated with cognitive impairment in PD.

In all these studies, the input to the learning algorithm consists of a limited set of features extracted from the saccade task, such as reaction time, latency, saccade duration, and saccade amplitude. Feeding the model with these predefined sets of values instead of the raw data helps to overcome the problem of overfitting and allows the use of simpler ML models. However, it also significantly limits the amount of information available in the biomedical signal, preventing the model from discovering new relevant features and utilizing these for subsequent classification. 
\subsection{Contribution}

In this paper, we present a novel machine learning approach to discriminate between PD and HC using eye-tracking data from a pro-/antisaccade experiment [\cite{waldthaler_vertical_2019}]. In contrast to previous studies, we focus on the simplest possible task the eye can perform, fixation. Instead of analyzing the saccades themselves, we base our analysis on short fixation time intervals of $\sim1.5\,s$ length, corresponding to the pre-saccade preparation phase. This approach is motivated by previous findings, which show that fixation data carries information about a subject's condition [\cite{tsitsi_fixation_2021}]. Furthermore, rather than defining hand-crafted features extracted from the eye-tracking data, we use the raw time series data containing the gaze position, which allows the model to discover features which are useful for the classification task.

We show that, with proper data preprocessing and hyperparameter tuning, state-of-the-art time series classification models trained on fixation eye-tracking data can successfully discriminate between PD and HC. We implement the InceptionTime and ROCKET algorithms, which achieve $56\%$ and $68\%$ accuracy, respectively, in classifying individual fixation trials, and $78\%$ and $88\%$ accuracy, respectively, in classifying subjects. 


Furthermore, we provide a comprehensive analysis of the ROCKET classifiers. Using a recently developed feature selection algorithm referred to as Sequential Feature Detachment (SFD) [\cite{uribarri_detach-rocket_2023}], we reduce the average number of parameters of the model by $92\%$ percent while improving the classification performance on the test set. Utilizing SFD, we reach accuracies of $73\%$ and $96\%$ at trial and subject level, respectively. This improvement in model performance underscore the potential of SFD for improving model generalization on challenging real-world datasets. We also demonstrate how the selected features can be used to interpret the model.

Our results suggest that the homogeneity of the fixation task minimizes subject identity fingerprints in the data, allowing the models to generalize well to unseen subjects. This makes the application of machine learning to eye-tracking data from gaze fixation a promising strategy for non-invasive diagnosis and staging of health conditions, not only in Parkinson's disease, but possibly also in other cognitive or motor disorders.


\section{Background}
\label{sec:background}
\subsection{Machine Learning for Time Series Classification}
Due to the increasing availability of computing power and the constant evolution of available algorithms, the field of Time Series Classification (TSC) is dominated by machine learning algorithms [\cite{ismail_fawaz_deep_2019, faouzi_time_2022}]. The more efficient and effective TSC algorithms are those that exploit the sequential nature of the data and are therefore able to identify features that depend on the temporal order of the input values [\cite{elman_finding_1990, uribarri_dynamical_2022}]. RNNs, LSTMs [\cite{hochreiter_long_1997}], and Shapelets [\cite{ye_time_2011}] are some traditional examples from this class of algorithms.

In recent years, a number of new state-of-the-art machine learning models have been developed that outperform traditional approaches in most tasks [\cite{ruiz_great_2021}]. Two of the leading models for Multivariate Time Series (MTS) classification are InceptionTime, a deep learning based method [\cite{ismail_fawaz_inceptiontime_2020}], and ROCKET, a transformation ensemble method [\cite{dempster_rocket_2020}]. A brief description of each of these models is presented in the following subsections. 

\subsection{InceptionTime}
InceptionTime was inspired by the Inception models, a lineage of multi-focal deep convolutional neural networks for image classification [\cite{szegedy_going_2015, szegedy_inception-v4_2017}]. The key building block of InceptionTime is the inception module. The inception module performs a set of parallel convolutions using kernels of varying size, thus attending to patterns of multiple time scales. Importantly, the inception module also features a bottleneck layer, which limits the number of parameters in the model and prevents overfitting when using small datasets. InceptionTime stacks multiple inception modules, producing a high-dimensional embedding which is fed to a fully connected layer for classification.



When compared to other popular methods for MTS classification, InceptionTime demonstrates superior performance on many different datasets [\cite{ruiz_great_2021}]. InceptionTime also exhibits impressive scaling characteristics with respect to time series length and the number of training samples [\cite{ismail_fawaz_inceptiontime_2020}].

\subsection{ROCKET and SFD}

ROCKET is a computationally inexpensive and accurate model for time series classification [\cite{dempster_rocket_2020}]. A massive number of non-trainable random convolutional kernels are applied to the input time series, transforming it into feature maps. The maximum value and proportion of positive values are computed for each feature map, producing the final attributes which are fed to a classifier. The authors propose to use a ridge regression classifier for small to medium sized datasets, and a linear fully connected layer trained with stochastic gradient descent for large datasets [\cite{dempster_rocket_2020}].

ROCKET offers performance comparable to state-of-the-art methods at a comparably diminutive compute budget [\cite{dempster_rocket_2020,ruiz_great_2021}]. Since the convolutional stage of ROCKET is non-trainable, only the linear classifier needs to be fit to the data. The low number of trainable parameters not only mitigate the risk of overfitting, but also allows the model to be trained on large datasets - out of reach for previous models [\cite{dempster_rocket_2020}].

Recently, a new methodology known as Sequential Feature Detachment (SFD) has been introduced for sequential selection and pruning of features in ROCKET models. This methodology leverages the linearity of ROCKETS's classifier to iteratively remove the least informative features. SFD significantly reduces the model's parameter count without compromising the test set accuracy. In most cases, it has been shown to improve model generalization [\cite{uribarri_detach-rocket_2023}].

\section{Cohort, Data Collection and Preprocessing}
\label{sec:data}


\subsection{Cohort Description and Data Collection}
\label{sec:cohort}
The dataset consists of data coming from 84 participants, 54 non-demented PD patients and 30 age-matched healthy controls. Participants were asked to perform horizontal prosaccade and antisaccade tasks. In the case of PD patients, the procedure was conducted on two occations where possible: first with patients on their regular dose of dopamine replacement therapy (levodopa) and then after a 12 hour withdrawal from their dopaminergic medication. We refer to these conditions as PD ON and PD OFF. The study was approved by the local ethics committee (2016/348-31/4) of Karolinska University Stockholm and all participants gave written informed consent. The data was anonymized for use in the present work.


In the prosaccade task, participants were asked to fixate a central point on the screen and then move their eyes as quickly as possible to a target that appeared to the left or right of the fixation point. In contrast, in the antisaccade task, participants were required to suppress the reflexive saccade and instead make a voluntary eye movement in the opposite direction.

Eye movements were recorded using a binocular head-mounted eye-tracker (EyeBrain T2, medical device with CE label for clinical use Class IIa, ISO 9001, ISO 13485) which tracks the pupil using near-infrared light. Data were acquired for both eyes at a sampling rate of 300 Hz. Stimuli were presented on a 22" widescreen at a distance of 60 cm. 

A more extensive description of the data collection, cohort, and experimental protocol can be found in a previous publication of ours [\cite{waldthaler_vertical_2019}]. 

Throughout the paper we will use the following nomenclature. A session refers to one recording event where data is collected from a subject during a pro- or antisaccade experiment. A session consists of a series of trials, i.e stimuli presented to the subject. A data point refers to a single measurement in time of the eye position and velocity.

\subsection{Data Sanitization and Filtering}
\label{sec:preprocessing}

\begin{figure*}[t]
  \centering 
  \includegraphics[width=\textwidth]{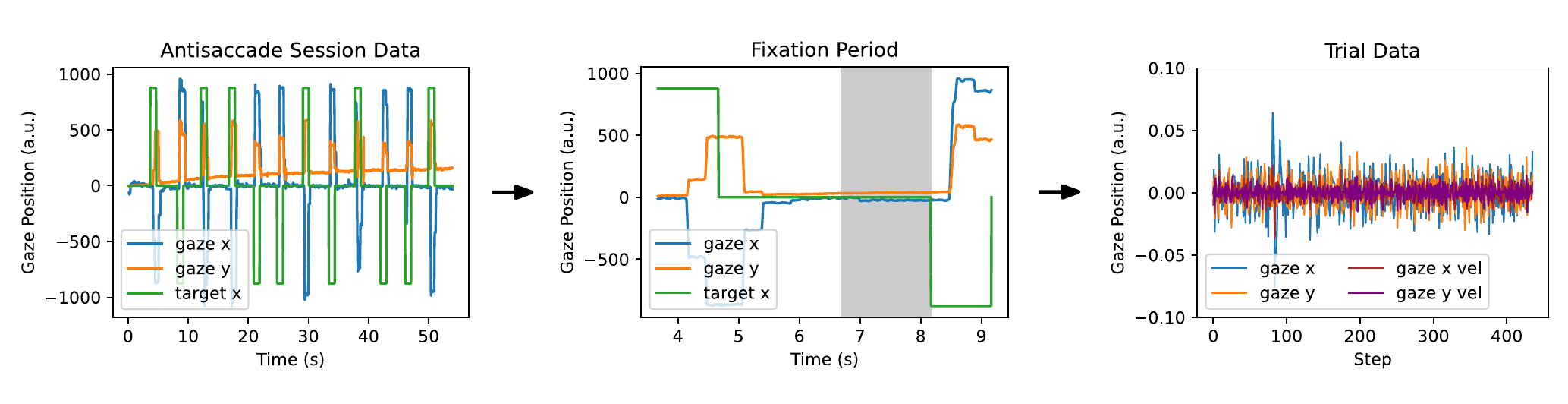} 
  \caption{Eye-tracking data preprocessing pipeline for the saccade experiments. Each session is composed of 12 consecutive prosaccades or antisaccades. The data was split into time intervals corresponding to $\sim1.5\,s$ before the target moves (gray area). A trial is then composed of the x and y positions and velocities during that time interval, high-pass filtered at 20Hz.}
  \label{fig:preprocessing} 
\end{figure*}

A summary of the preprocessing pipeline is shown in figure \ref{fig:preprocessing}. A detail description of the data sanitization and splitting of sessions into trials can be found in Appendix \ref{apd:preprocessing}.

In preprocessing the eye tracking data, it is crucial to remove any traces of participants' head movements. For PD patients, these traces might reveal information about their distinctive resting tremor, potentially influencing the training of the classifier. To counter this, we applied an eighth-order Butterworth high-pass filter to the time series under study. The methodology for choosing the filter's cutoff frequency is presented in section \ref{sec:cutoff}.

\subsection{Data Splitting and Upsampling}
\label{sec:data_split}

Data splitting is a pivotal step of the process, due to the high variability in clinical data and the small number of subjects. If data from the same subject is present in both training and test data, data leakage might allow the model to correctly classify the test data by learning subject specific features during training. In that case, the test set performance does not reflect the model's ability to generalize to unseen subjects. To counteract data leakage, we performed the train-val-test split $(880, 264, 440)$ such that each split was mutually exclusive in terms of subjects.

To ensure that the statistics on the validation and test splits accurately reflect the performance for all classes, we performed a stratified train-val-test split such that the class distribution is approximately equal for all splits. For training, PD OFF and PD ON data was pooled and the classification was conducted as a binary problem: PD versus HC.


The class balance in the dataset is also of critical importance, since an imbalanced class distribution might incentivize the model to overwhelmingly predict the majority class. The dataset we used suffered from a class imbalance, having less data from healthy controls than PD patients. During training, we upsampled the minority class in order to achieve an even class distribution. The up-sampling was done during data loading by weighting the probability of selecting a trial from class $C$ by $1/N_c$, where $N_c$ was the number of trials belonging to class $C$.

\section{Methods}
\label{sec:methods}

\subsection{Metrics and Evaluation Criteria}
\label{sec:metrics}

We report the accuracy and unweighted F1-score (uF1) for our experiments. The F1-score represents the harmonic mean of the precision and recall, also referred to as the positive predictive value and sensitivity. Because our validation and test set have imbalanced class distributions, we report the unweighted F1-score, meaning that the precision and recall of both classes are averaged with equal weight. The accuracy is reported to provide an interpretable metric which easily conveys the performance of the model. The uF1-score is used to compare models and hyperparameter configurations, due to its expressiveness and robustness to class imbalance. 

\subsection{Model Implementation}
A comprehensive overview of the InceptionTime and ROCKET model implementations can be found in Appendix \ref{apd:model}. For pruning the ROCKET model with SFD, we utilized the implementation provided in the official repository of the paper \url{https://github.com/gon-uri/detach_rocket} [\cite{uribarri_detach-rocket_2023}].

\subsection{Prediction Aggregation for Subject Classification}
\label{sec:aggregation}

\begin{figure*}[ht]
  \centering 
  \includegraphics[width=0.75\textwidth]{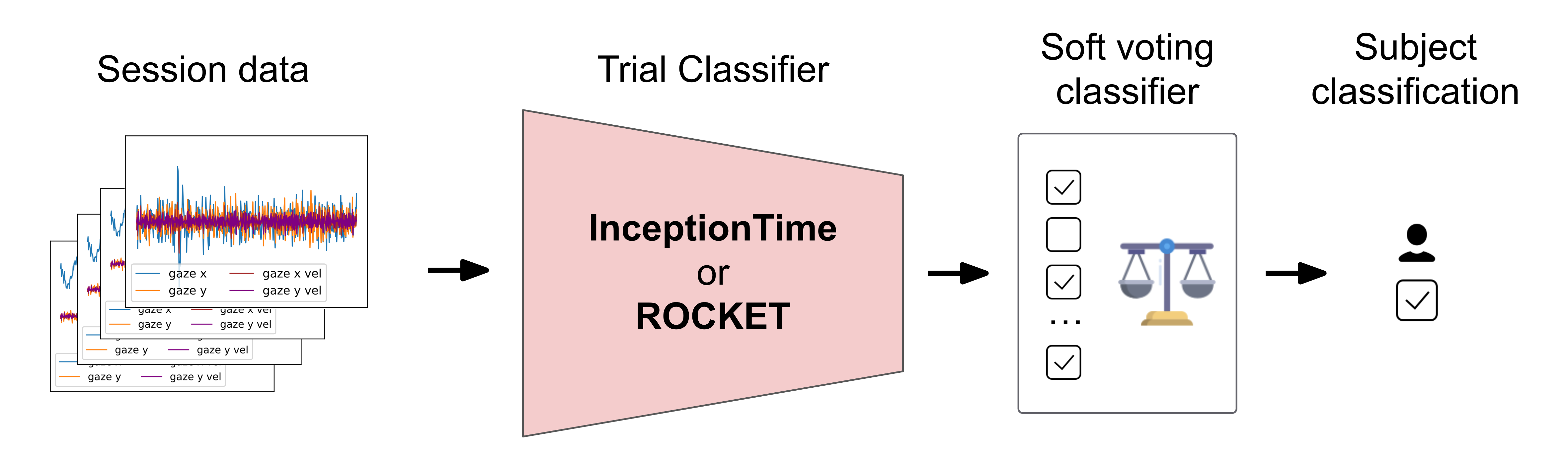} 
  \caption{Schematics of the end-to-end model used for subject level classification.}
  \label{fig:model} 
\end{figure*}

In order to make predictions on subject-level, we aggregated the trial-level predictions for each subject and implemented a soft voting classifier. After making predictions on all trials performed by one subject, the trial-level probability scores were averaged to produce a subject-level probability score. By thresholding the subject-level probability score, we made a binary prediction of whether the subject was an HC or PD patient. This is one of the simplest possible ways in which this aggregation can be done. Other alternatives involve training a model to integrate the trial level predictions. However, the small number of subjects in the training set makes employing such an approach very challenging. In figure \ref{fig:model} we present the schematics of the end-to-end model used for subject level classification.

\section{Results} 
\label{sec:results}

\subsection{Filter Cutoff Frequency Selection}
\label{sec:cutoff}

To determine an appropriate high-pass cutoff frequency, we assessed the performance of a ROCKET classifier trained on eye tracking data that was progressively filtered with increasing cutoff frequencies. Our goal was to ensure that the signal's frequency region corresponding to the primary PD tremor and its first harmonic was removed. These regions have reported values between $4\ \text{Hz}-7\ \text{Hz}$ and $8\ \text{Hz}-14\ \text{Hz}$, respectively [\cite{chan_motion_2022}, \cite{raethjen_cortical_2009}]. At the same time, it was essential to preserve the eye movement information relevant for the classification task.

\begin{figure}[t]
  \centering 
  \includegraphics[width=0.45\textwidth]{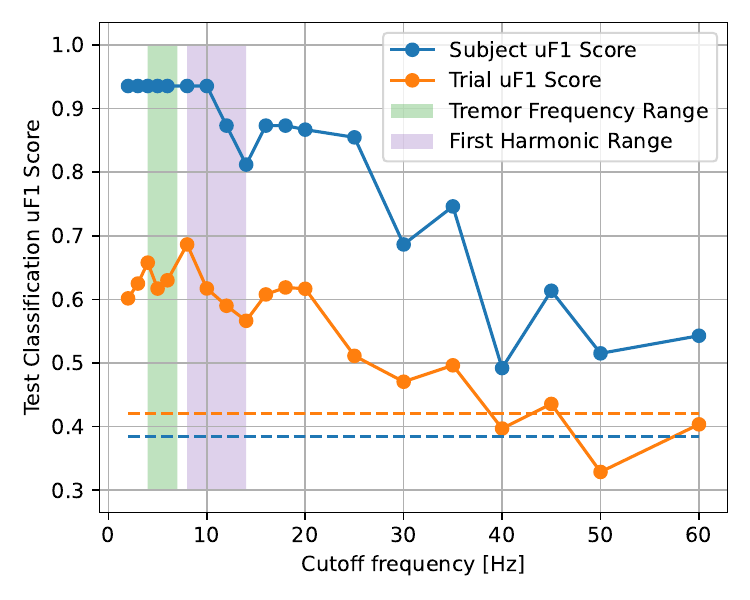} 
  \caption{Performance a ROCKET classifier trained on filtered data as a function of the cutoff frequency. Performance is reported as uF1 both at trial and subject level. Dashed lines correspond to the uF1 score of a benchmark model that consistently predicts the predominant class. Frequency regions corresponding to resting PD tremor and its first harmonic are highlighted.}
  \label{fig:cutoff} 
\end{figure}

Figure \ref{fig:cutoff} illustrates the relationship between the cutoff frequency and the model's classification performance for the test set at both trial and subject levels. The figure reveals a region (15Hz to 25Hz) where both the tremor and its harmonic have been eliminated, yet the classifier maintains significant predictive power (surpassing a benchmark model predicting only the predominant class). Notice that when the cutoff frequency exceeds 35Hz, the model's performance noticeably declines, suggesting the loss of critical eye movement data. Based on our analysis, we opted for a conservative 20Hz as the cutoff frequency for the high-pass filter.

\subsection{Model Hyperparameter Search}

To ensure a fair comparison between models, we performed a hyperparameter search for InceptionTime and ROCKET. The hyperparameter configurations were evaluated by the maximum uF1-scores averaged over a stratified 5-fold cross validation, where each fold was mutually exclusive in terms of subjects. (see section \ref{sec:data_split})

\paragraph{Model 1: InceptionTime}

For InceptionTime, we searched values for the learning rate, batch size, number of inception time modules, and hidden dimensionality. We estimated the performance after 100 epochs of training, as this has been demonstrated to accurately reflect the final performance of deep neural networks [\cite{domhan_speeding_2015}]. First we performed a grid search on learning rate and batch size, which are intimately related [\cite{goyal_accurate_2018}]. Then we continued with a grid search on the number of inception time modules and hidden dimensionality, accounting for the model size. The numerical results from the hyperparameter search with InceptionTime are presented in table \ref{tab:it_hyperparam_search} of Appendix \ref{apd:hyperparam}. A relatively large batch size and learning rate seems to offer the best performance, while a depth of 4 inception modules appears to be optimal for our input sequence length. Also, increasing the hidden dimensionality, and thus the expressiveness of the model, seems to boost performance.

\paragraph{Model 2: ROCKET}
In the case of ROCKET, the hyperparameter search included the number of kernels in the transformation stage and the ridge parameter, which controls the amount of L2 regularization in the classifier. The results from the hyperparameter search performed on ROCKET are presented in table \ref{tab:rocket_param_search} of Appendix \ref{apd:hyperparam}. In agreement with the original authors, we find that the optimal number of kernels is the default number of 10\,000 [\cite{dempster_rocket_2020}].

\subsection{Classification Results}
\begin{table*}[h]
\centering
\begin{tabular}{@{}lcccc@{}}
\multicolumn{1}{c}{} & \multicolumn{2}{c}{Trial} & \multicolumn{2}{c}{Subject} \\
Model                & uF1-Score    & Accuracy   & uF1-Score     & Accuracy    \\ \midrule
InceptionTime        & $0.52\pm0.02$       & $55.73\%\pm2.84\%$    & $0.74\pm0.04$        & $77.50\%\pm5.59\%$    \\
ROCKET               & $0.63\pm0.02$       & $68.04\%\pm2.23\%$    & $0.86\pm0.04$        & $87.50\%\pm4.42\%$ \\ 
Detach-ROCKET  & $\textbf{0.66}\pm0.02$       & $\textbf{73.46\%}\pm0.85\%$    & $\textbf{0.96}\pm0.04$        & $\textbf{96.25\%}\pm3.42\%$ \\ \midrule
\cite{brien_classification_2023}         & $\ $       & $\ $      & $\ $        & $82\%\pm6.7\%$      \\  \bottomrule
\end{tabular}    \\
\caption{The first three entries show the classification performance for InceptionTime, ROCKET and Detach-RCOKET. We report the mean and standard deviation over five training runs with different random initializations. The last row shows the result of a previous study where the median over a cross-validation process is reported.}  

\label{tab:main}
\end{table*}

\noindent Table \ref{tab:main} presents the results for trial and subject classification tasks for both models. To obtain an estimate of the variance in the performance of the predictions, training was performed 5 times for each model with different random initializations. ROCKET performs better than InceptionTime at trial and subject level, achieving remarkable accuracy on subject classification. For comparison, the table \ref{tab:main} also shows the results obtained in [\cite{brien_classification_2023}]. Note that this study used 45 features for classification, many of which are from sources not included as input to our models (e.g. error rates, anticipation rate, pupil size, reaction time). 

\subsection{Reduced ROCKET and Interpretability}

\paragraph{Feature Selection with SFD.}
We applied the Sequential Feature Detachment methodology to the reported ROCKET models [\cite{uribarri_detach-rocket_2023}]. With the default size-performance tradeoff hyperparameter $(c=0.1)$, the reduced models retained on average merely $8.46\%\pm7.93\%$ of the original features. The test set performance of these models is presented in Table \ref{tab:main} under the label Detach-ROCKET. When compared to the full models, the reduced models perform notably better at both trial and subject level.



\begin{figure}[t]
  \centering 
  \includegraphics[width=0.4\textwidth]{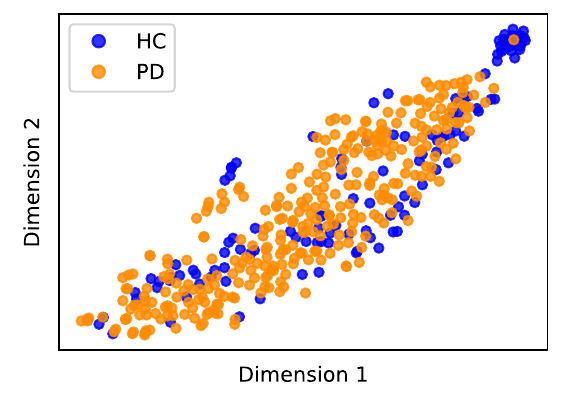} 
  \caption{Low-dimensional representation of the test set using the best Detach-ROCKET model (retaining $2.25\%$ of features). We present the separation between healthy controls (blue) and patients with Parkinson's disease (orange).}
  \label{fig:latent} 
\end{figure}

\paragraph{Interpretability.}
A reduced ROCKET model facilitates interpretability analyses that may be impractical or unattainable with the full model. For instance, as only the informative features are retained, we can examine a low-dimensional representation of this feature space using t-SNE [\cite{van_der_maaten_visualizing_2008}]. Figure \ref{fig:latent} depicts the distribution of test trials in a reduced 2D feature space for our top-performing Detach-ROCKET model, revealing two types of HC trials: one that is more distinct, and another that is more difficult to distinguish from PD trials.




\subsection{Attribute Analysis}
We investigated potential variations in the classifier's predictive capabilities based on different trial attributes. To achieve this, we evaluated five ROCKET models, each initialized with a unique random seed, and then calculated the average and standard deviation of the classification accuracy for each trial type.

\paragraph{Prosaccades vs. Antisaccade.}
We found that the accuracy for trials preceding a prosaccade is $71.55\% \pm 2.02\%$, while for trials preceding an antisaccade it is $64.54\% \pm 2.02\%$, suggesting a notable difference in difficulty for the classifier in identifying the subject's condition.

\paragraph{ON Medication vs. OFF Medication.}
On the contrary, we found no significant difference in predictive ability between ON and OFF medication trials. The accuracy for ON medication trials is $66.06\% \pm 2.75\%$, while for OFF medication trials it is $64.61\% \pm 1.15\%$.



\section{Discussion} 
\label{sec:discussion}


It is remarkable that a machine learning model trained on data from only 47 subjects can generalize to unseen subjects. This is even more impressive considering that each subject's trials share the same label, intertwining the target class variability with the subjects’ identities. The model's success in generalizing shows that it is effective at learning features relevant to the target classes and less prone to overfitting to subject-specific features from the training data.

This success suggests that eye-tracking data from short fixation intervals carry relevant information about health status, while at the same time expressing low intersubject variability - low enough that even deep learning models can effectively learn disease features without overfitting subjects. This low variability likely stems from the simplicity and homogeneity of the fixation task, which does not involve any large voluntary eye movements.

Our results align with previous research which indicates that subtle involuntary eye movements made during target fixation convey significant health-related information [\cite{tsitsi_fixation_2021}]. The combination of carrying health status information and having low intersubject variability makes fixation data a promising candidate for data-driven biomarker discovery, especially in cases where symptoms include visual impairment, such as Parkinson's disease.


The main limitation of this work is its reliance on data from a single experiment. This not only limits the number of subjects available, but also creates homogeneity in the recording conditions. Future work expanding the dataset to other saccade experiments could shed light on how the results generalize across experimental conditions. 

Concerning experimental conditions, another interesting discussion arises from the comparison between prosaccade and antisaccade trials. The antisaccade task involves inhibiting the intuitive gaze reflex to move in the direction of the target and is therefore known to be more cognitively demanding than the prosaccade task. Recent studies have shown that when brain activity during the preparation phase is examined using magnetoencephalography, differences between PD patients and HC are significant for the antisaccade task but not for the prosaccade task [\cite{waldthaler_neural_2022}]. If we hypothesize that eye movements carry information about the subjects' cognitive state, it is natural to ask whether trials corresponding to fixation in preparation for the antisaccade task have different classification accuracy than those in preparation for the prosaccade task. Our result suggests that preparatory fixations preceding a prosaccade are more informative than those preceding an antisaccade, the opposite of what has been observed in brain data. However, given the limited number of patients in the test set, these results should be considered as indicative rather than conclusive.


\section{Conclusion} 
\label{sec:conclusion}


In this study, we investigated the use of deep learning algorithms to discriminate between healthy controls and Parkinson's diseased patients using eye-tracking data from saccade experiments. In contrast to previous approaches, instead of using hand-crafted features on the saccade data, we analyzed the continuous eye-movements during $\sim 1.5\ s$ time intervals corresponding to the pre-task fixation period. 

We evaluated two state-of-the-art machine learning models for the task, InceptionTime and ROCKET. Our experiments concluded that ROCKET performed better, achieving accuracies of $68\%$ for trial classification and $88\%$ for subject classification on data from unseen subjects. Using SFD, we were able to drastically reduce the size of the ROCKET model, keeping on average only $8\%$ of parameters. Remarkably, in doing so we improved the accuracy at trial and subject level to $73\%$ and $96\%$, respectively. Applying SFD also allowed us to explore the latent representation of classes using only relevant features, thereby improving the interpretability of the model.



Our results show that short intervals of fixation data can be used to correctly classify unseen subjects as PD or HC. We speculate that the simplicity of gaze fixation reduces the amount of subject-specific features in the data, compared to more elaborate tasks where the heterogeneity of the subjects leads to subject-specific responses. Importantly, we show that the small involuntary eye movements in fixation data carry relevant information about a subject's health status. This motivates further research utilizing fixation data, and supports the hypothesis that eye-tracking devices are valuable tools for developing new diagnostic and prognostic protocols for Parkinson's disease.



\bibliography{zotero_references_gon}

\begin{thebibliography}{31}
\providecommand{\natexlab}[1]{#1}
\providecommand{\url}[1]{\texttt{#1}}
\expandafter\ifx\csname urlstyle\endcsname\relax
  \providecommand{\doi}[1]{doi: #1}\else
  \providecommand{\doi}{doi: \begingroup \urlstyle{rm}\Url}\fi

\bibitem[Anderson and MacAskill(2013)]{anderson_eye_2013}
Tim~J. Anderson and Michael~R. MacAskill.
\newblock Eye movements in patients with neurodegenerative disorders.
\newblock \emph{Nature Reviews Neurology}, 9\penalty0 (2):\penalty0 74--85,
  February 2013.
\newblock ISSN 1759-4766.
\newblock \doi{10.1038/nrneurol.2012.273}.
\newblock URL \url{https://www.nature.com/articles/nrneurol.2012.273}.

\bibitem[Armstrong(2017)]{armstrong_visual_2017}
Richard~A. Armstrong.
\newblock Visual {Dysfunction} in {Parkinson}'s {Disease}.
\newblock In \emph{International {Review} of {Neurobiology}}, volume 134, pages
  921--946. Elsevier, 2017.
\newblock ISBN 978-0-12-812603-5.
\newblock URL
  \url{https://linkinghub.elsevier.com/retrieve/pii/S0074774217300375}.

\bibitem[Brien et~al.(2023)Brien, Riek, Yep, Huang, Coe, Areshenkoff, Grimes,
  Jog, Lang, Marras, Masellis, McLaughlin, Peltsch, Roberts, Tan, Beaton, Lou,
  Swartz, and Munoz]{brien_classification_2023}
Donald~C. Brien, Heidi~C. Riek, Rachel Yep, Jeff Huang, Brian Coe, Corson
  Areshenkoff, David Grimes, Mandar Jog, Anthony Lang, Connie Marras, Mario
  Masellis, Paula McLaughlin, Alicia Peltsch, Angela Roberts, Brian Tan, Derek
  Beaton, Wendy Lou, Richard Swartz, and Douglas~P. Munoz.
\newblock Classification and staging of {Parkinson}'s disease using video-based
  eye tracking.
\newblock \emph{Parkinsonism \& Related Disorders}, 110:\penalty0 105316, May
  2023.
\newblock ISSN 13538020.
\newblock \doi{10.1016/j.parkreldis.2023.105316}.
\newblock URL
  \url{https://linkinghub.elsevier.com/retrieve/pii/S1353802023000391}.

\bibitem[Chan et~al.(2022)Chan, Ripin, Halim, Arifin, Yahya, Eow, Tan, Hor, and
  Wong]{chan_motion_2022}
Ping~Yi Chan, Zaidi~Mohd Ripin, Sanihah~Abdul Halim, Wan~Nor Arifin,
  Ahmad~Shukri Yahya, Gaik~Bee Eow, Kenny Tan, Jyh~Yung Hor, and Chee~Keong
  Wong.
\newblock Motion characteristics of subclinical tremors in {Parkinson}’s
  disease and normal subjects.
\newblock \emph{Scientific Reports}, 12\penalty0 (1):\penalty0 4021, March
  2022.
\newblock ISSN 2045-2322.
\newblock \doi{10.1038/s41598-022-07957-z}.
\newblock URL \url{https://www.nature.com/articles/s41598-022-07957-z}.

\bibitem[Dempster et~al.(2020)Dempster, Petitjean, and
  Webb]{dempster_rocket_2020}
Angus Dempster, François Petitjean, and Geoffrey~I. Webb.
\newblock {ROCKET}: exceptionally fast and accurate time series classification
  using random convolutional kernels.
\newblock \emph{Data Mining and Knowledge Discovery}, 34\penalty0 (5):\penalty0
  1454--1495, September 2020.
\newblock ISSN 1573-756X.
\newblock \doi{10.1007/s10618-020-00701-z}.
\newblock URL \url{https://doi.org/10.1007/s10618-020-00701-z}.

\bibitem[Domhan et~al.(2015)Domhan, Springenberg, and
  Hutter]{domhan_speeding_2015}
Tobias Domhan, Jost~Tobias Springenberg, and Frank Hutter.
\newblock Speeding up {Automatic} {Hyperparameter} {Optimization} of {Deep}
  {Neural} {Networks} by {Extrapolation} of {Learning} {Curves}.
\newblock In \emph{Proceedings of the 24th {International} {Conference} on
  {Artificial} {Intelligence}}, {IJCAI}'15, pages 3460--3468. AAAI Press, 2015.
\newblock ISBN 978-1-57735-738-4.

\bibitem[Elman(1990)]{elman_finding_1990}
Jeffrey~L. Elman.
\newblock Finding {Structure} in {Time}.
\newblock \emph{Cognitive Science}, 14\penalty0 (2):\penalty0 179--211, March
  1990.
\newblock ISSN 03640213.
\newblock \doi{10.1207/s15516709cog1402_1}.
\newblock URL \url{http://doi.wiley.com/10.1207/s15516709cog1402_1}.

\bibitem[Faouzi(2022)]{faouzi_time_2022}
Johann Faouzi.
\newblock Time {Series} {Classification}: {A} review of {Algorithms} and
  {Implementations}.
\newblock In \emph{Machine {Learning} ({Emerging} {Trends} and
  {Applications})}. Proud Pen, 2022.
\newblock URL \url{https://inria.hal.science/hal-03558165}.

\bibitem[Goyal et~al.(2018)Goyal, Dollár, Girshick, Noordhuis, Wesolowski,
  Kyrola, Tulloch, Jia, and He]{goyal_accurate_2018}
Priya Goyal, Piotr Dollár, Ross Girshick, Pieter Noordhuis, Lukasz Wesolowski,
  Aapo Kyrola, Andrew Tulloch, Yangqing Jia, and Kaiming He.
\newblock Accurate, {Large} {Minibatch} {SGD}: {Training} {ImageNet} in 1
  {Hour}, April 2018.
\newblock URL \url{http://arxiv.org/abs/1706.02677}.

\bibitem[Hochreiter and Schmidhuber(1997)]{hochreiter_long_1997}
Sepp Hochreiter and Jürgen Schmidhuber.
\newblock Long {Short}-{Term} {Memory}.
\newblock \emph{Neural Computation}, 9\penalty0 (8):\penalty0 1735--1780,
  November 1997.
\newblock ISSN 0899-7667.
\newblock \doi{10.1162/neco.1997.9.8.1735}.

\bibitem[Ismail~Fawaz et~al.(2019)Ismail~Fawaz, Forestier, Weber, Idoumghar,
  and Muller]{ismail_fawaz_deep_2019}
Hassan Ismail~Fawaz, Germain Forestier, Jonathan Weber, Lhassane Idoumghar, and
  Pierre-Alain Muller.
\newblock Deep learning for time series classification: a review.
\newblock \emph{Data Mining and Knowledge Discovery}, 33\penalty0 (4):\penalty0
  917--963, July 2019.
\newblock ISSN 1573-756X.
\newblock \doi{10.1007/s10618-019-00619-1}.
\newblock URL \url{https://doi.org/10.1007/s10618-019-00619-1}.

\bibitem[Ismail~Fawaz et~al.(2020)Ismail~Fawaz, Lucas, Forestier, Pelletier,
  Schmidt, Weber, Webb, Idoumghar, Muller, and
  Petitjean]{ismail_fawaz_inceptiontime_2020}
Hassan Ismail~Fawaz, Benjamin Lucas, Germain Forestier, Charlotte Pelletier,
  Daniel~F. Schmidt, Jonathan Weber, Geoffrey~I. Webb, Lhassane Idoumghar,
  Pierre-Alain Muller, and François Petitjean.
\newblock {InceptionTime}: {Finding} {AlexNet} for time series classification.
\newblock \emph{Data Mining and Knowledge Discovery}, 34\penalty0 (6):\penalty0
  1936--1962, November 2020.
\newblock ISSN 1573-756X.
\newblock \doi{10.1007/s10618-020-00710-y}.
\newblock URL \url{https://doi.org/10.1007/s10618-020-00710-y}.

\bibitem[Kelly et~al.(2019)Kelly, Karthikesalingam, Suleyman, Corrado, and
  King]{kelly_key_2019}
Christopher~J. Kelly, Alan Karthikesalingam, Mustafa Suleyman, Greg Corrado,
  and Dominic King.
\newblock Key challenges for delivering clinical impact with artificial
  intelligence.
\newblock \emph{BMC Medicine}, 17\penalty0 (1):\penalty0 195, October 2019.
\newblock ISSN 1741-7015.
\newblock \doi{10.1186/s12916-019-1426-2}.
\newblock URL \url{https://doi.org/10.1186/s12916-019-1426-2}.

\bibitem[Przybyszewski et~al.(2016)Przybyszewski, Kon, Szlufik, Szymanski,
  Habela, and Koziorowski]{przybyszewski_multimodal_2016}
Andrzej~W. Przybyszewski, Mark Kon, Stanislaw Szlufik, Artur Szymanski, Piotr
  Habela, and Dariusz~M. Koziorowski.
\newblock Multimodal {Learning} and {Intelligent} {Prediction} of {Symptom}
  {Development} in {Individual} {Parkinson}’s {Patients}.
\newblock \emph{Sensors}, 16\penalty0 (9):\penalty0 1498, September 2016.
\newblock ISSN 1424-8220.
\newblock \doi{10.3390/s16091498}.
\newblock URL \url{https://www.mdpi.com/1424-8220/16/9/1498}.

\bibitem[Przybyszewski et~al.(2023)Przybyszewski, Śledzianowski, Chudzik,
  Szlufik, and Koziorowski]{przybyszewski_machine_2023}
Andrzej~W. Przybyszewski, Albert Śledzianowski, Artur Chudzik, Stanisław
  Szlufik, and Dariusz Koziorowski.
\newblock Machine {Learning} and {Eye} {Movements} {Give} {Insights} into
  {Neurodegenerative} {Disease} {Mechanisms}.
\newblock \emph{Sensors}, 23\penalty0 (4):\penalty0 2145, January 2023.
\newblock ISSN 1424-8220.
\newblock \doi{10.3390/s23042145}.
\newblock URL \url{https://www.mdpi.com/1424-8220/23/4/2145}.

\bibitem[Raethjen et~al.(2009)Raethjen, Govindan, Muthuraman, Kopper, Volkmann,
  and Deuschl]{raethjen_cortical_2009}
Jan Raethjen, R.B. Govindan, M.~Muthuraman, Florian Kopper, Jens Volkmann, and
  Günther Deuschl.
\newblock Cortical correlates of the basic and first harmonic frequency of
  {Parkinsonian} tremor.
\newblock \emph{Clinical Neurophysiology}, 120\penalty0 (10):\penalty0
  1866--1872, October 2009.
\newblock ISSN 13882457.
\newblock \doi{10.1016/j.clinph.2009.06.028}.
\newblock URL
  \url{https://linkinghub.elsevier.com/retrieve/pii/S138824570900488X}.

\bibitem[Rajpurkar et~al.(2022)Rajpurkar, Chen, Banerjee, and
  Topol]{rajpurkar_ai_2022}
Pranav Rajpurkar, Emma Chen, Oishi Banerjee, and Eric~J. Topol.
\newblock {AI} in health and medicine.
\newblock \emph{Nature Medicine}, 28\penalty0 (1):\penalty0 31--38, January
  2022.
\newblock ISSN 1546-170X.
\newblock \doi{10.1038/s41591-021-01614-0}.
\newblock URL \url{https://www.nature.com/articles/s41591-021-01614-0}.

\bibitem[Ruiz et~al.(2021)Ruiz, Flynn, Large, Middlehurst, and
  Bagnall]{ruiz_great_2021}
Alejandro~Pasos Ruiz, Michael Flynn, James Large, Matthew Middlehurst, and
  Anthony Bagnall.
\newblock The great multivariate time series classification bake off: a review
  and experimental evaluation of recent algorithmic advances.
\newblock \emph{Data Mining and Knowledge Discovery}, 35\penalty0 (2):\penalty0
  401--449, March 2021.
\newblock ISSN 1384-5810, 1573-756X.
\newblock \doi{10.1007/s10618-020-00727-3}.
\newblock URL \url{http://link.springer.com/10.1007/s10618-020-00727-3}.

\bibitem[Sledzianowski et~al.(2019)Sledzianowski, Szymanski, Drabik, Szlufik,
  Koziorowski, and Przybyszewski]{sledzianowski_measurements_2019}
Albert Sledzianowski, Artur Szymanski, Aldona Drabik, Stanisław Szlufik,
  Dariusz~M. Koziorowski, and Andrzej~W. Przybyszewski.
\newblock Measurements of {Antisaccades} {Parameters} {Can} {Improve} the
  {Prediction} of {Parkinson}’s {Disease} {Progression}.
\newblock In \emph{Intelligent {Information} and {Database} {Systems}}, volume
  11432, pages 602--614. Springer International Publishing, Cham, 2019.
\newblock ISBN 978-3-030-14801-0 978-3-030-14802-7.
\newblock URL \url{http://link.springer.com/10.1007/978-3-030-14802-7_52}.

\bibitem[Stuart et~al.(2019)Stuart, Lawson, Yarnall, Nell, Alcock, Duncan,
  Khoo, Barker, Rochester, Burn, and Group]{stuart_pro-saccades_2019}
Samuel Stuart, Rachael~A. Lawson, Alison~J. Yarnall, Jeremy Nell, Lisa Alcock,
  Gordon~W. Duncan, Tien~K. Khoo, Roger~A. Barker, Lynn Rochester, David~J.
  Burn, and on~behalf of the ICICLE-PD~study Group.
\newblock Pro-{Saccades} {Predict} {Cognitive} {Decline} in {Parkinson}'s
  {Disease}: {ICICLE}-{PD}.
\newblock \emph{Movement Disorders}, 34\penalty0 (11):\penalty0 1690--1698,
  2019.
\newblock ISSN 1531-8257.
\newblock \doi{10.1002/mds.27813}.
\newblock URL \url{https://onlinelibrary.wiley.com/doi/abs/10.1002/mds.27813}.

\bibitem[Szegedy et~al.(2015)Szegedy, {Wei Liu}, {Yangqing Jia}, Sermanet,
  Reed, Anguelov, Erhan, Vanhoucke, and Rabinovich]{szegedy_going_2015}
Christian Szegedy, {Wei Liu}, {Yangqing Jia}, Pierre Sermanet, Scott Reed,
  Dragomir Anguelov, Dumitru Erhan, Vincent Vanhoucke, and Andrew Rabinovich.
\newblock Going {Deeper} with {Convolutions}.
\newblock In \emph{2015 {IEEE} {Conference} on {Computer} {Vision} and
  {Pattern} {Recognition} ({CVPR})}, pages 1--9, Boston, MA, USA, June 2015.
  IEEE.
\newblock ISBN 978-1-4673-6964-0.
\newblock \doi{10.1109/CVPR.2015.7298594}.
\newblock URL \url{http://ieeexplore.ieee.org/document/7298594/}.

\bibitem[Szegedy et~al.(2017)Szegedy, Ioffe, Vanhoucke, and
  Alemi]{szegedy_inception-v4_2017}
Christian Szegedy, Sergey Ioffe, Vincent Vanhoucke, and Alexander Alemi.
\newblock Inception-v4, {Inception}-{ResNet} and the {Impact} of {Residual}
  {Connections} on {Learning}.
\newblock \emph{Proceedings of the AAAI Conference on Artificial Intelligence},
  31\penalty0 (1), February 2017.
\newblock ISSN 2374-3468.
\newblock \doi{10.1609/aaai.v31i1.11231}.
\newblock URL \url{https://ojs.aaai.org/index.php/AAAI/article/view/11231}.

\bibitem[Termsarasab et~al.(2015)Termsarasab, Thammongkolchai, Rucker, and
  Frucht]{termsarasab_diagnostic_2015}
Pichet Termsarasab, Thananan Thammongkolchai, Janet~C. Rucker, and Steven~J.
  Frucht.
\newblock The diagnostic value of saccades in movement disorder patients: a
  practical guide and review.
\newblock \emph{Journal of Clinical Movement Disorders}, 2\penalty0
  (1):\penalty0 14, October 2015.
\newblock ISSN 2054-7072.
\newblock \doi{10.1186/s40734-015-0025-4}.
\newblock URL \url{https://doi.org/10.1186/s40734-015-0025-4}.

\bibitem[Tsitsi et~al.(2021)Tsitsi, Benfatto, Seimyr, Larsson, Svenningsson,
  and Markaki]{tsitsi_fixation_2021}
Panagiota Tsitsi, Mattias~Nilsson Benfatto, Gustaf~Öqvist Seimyr, Olof
  Larsson, Per Svenningsson, and Ioanna Markaki.
\newblock Fixation {Duration} and {Pupil} {Size} as {Diagnostic} {Tools} in
  {Parkinson}’s {Disease}.
\newblock \emph{Journal of Parkinson's Disease}, 11\penalty0 (2):\penalty0
  865--875, January 2021.
\newblock ISSN 1877-7171.
\newblock \doi{10.3233/JPD-202427}.
\newblock URL
  \url{https://content.iospress.com/articles/journal-of-parkinsons-disease/jpd202427}.

\bibitem[Uribarri and Mindlin(2022)]{uribarri_dynamical_2022}
Gonzalo Uribarri and Gabriel~B. Mindlin.
\newblock Dynamical time series embeddings in recurrent neural networks.
\newblock \emph{Chaos, Solitons \& Fractals}, 154:\penalty0 111612, January
  2022.
\newblock ISSN 0960-0779.
\newblock \doi{10.1016/j.chaos.2021.111612}.
\newblock URL
  \url{https://www.sciencedirect.com/science/article/pii/S0960077921009668}.

\bibitem[Uribarri et~al.(2023)Uribarri, Barone, Ansuini, and
  Fransén]{uribarri_detach-rocket_2023}
Gonzalo Uribarri, Federico Barone, Alessio Ansuini, and Erik Fransén.
\newblock Detach-rocket: Sequential feature selection for time series
  classification with random convolutional kernels.
\newblock 2023.
\newblock \doi{10.48550/arXiv.2309.14518}.
\newblock URL \url{https://arxiv.org/abs/2309.14518}.

\bibitem[Van~der Maaten and Hinton(2008)]{van_der_maaten_visualizing_2008}
Laurens Van~der Maaten and Geoffrey Hinton.
\newblock Visualizing data using t-{SNE}.
\newblock \emph{Journal of machine learning research}, 9\penalty0 (11), 2008.

\bibitem[Waldthaler et~al.(2019)Waldthaler, Tsitsi, and
  Svenningsson]{waldthaler_vertical_2019}
Josefine Waldthaler, Panagiota Tsitsi, and Per Svenningsson.
\newblock Vertical saccades and antisaccades: complementary markers for motor
  and cognitive impairment in {Parkinson}’s disease.
\newblock \emph{npj Parkinson's Disease}, 5\penalty0 (1):\penalty0 1--6, June
  2019.
\newblock ISSN 2373-8057.
\newblock \doi{10.1038/s41531-019-0083-7}.
\newblock URL \url{https://www.nature.com/articles/s41531-019-0083-7}.

\bibitem[Waldthaler et~al.(2022)Waldthaler, Vinding, Eriksson, Svenningsson,
  and Lundqvist]{waldthaler_neural_2022}
Josefine Waldthaler, Mikkel~C. Vinding, Allison Eriksson, Per Svenningsson, and
  Daniel Lundqvist.
\newblock Neural correlates of impaired response inhibition in the antisaccade
  task in {Parkinson}’s disease.
\newblock \emph{Behavioural Brain Research}, 422:\penalty0 113763, March 2022.
\newblock ISSN 0166-4328.
\newblock \doi{10.1016/j.bbr.2022.113763}.
\newblock URL
  \url{https://www.sciencedirect.com/science/article/pii/S0166432822000316}.

\bibitem[Waldthaler et~al.(2023)Waldthaler, Stock, Krüger‐Zechlin, Deeb, and
  Timmermann]{waldthaler_cluster_2023}
Josefine Waldthaler, Lena Stock, Charlotte Krüger‐Zechlin, Zain Deeb, and
  Lars Timmermann.
\newblock Cluster analysis reveals distinct patterns of saccade impairment and
  their relation to cognitive profiles in {Parkinson}'s disease.
\newblock \emph{Journal of Neuropsychology}, 17\penalty0 (2):\penalty0
  251--263, June 2023.
\newblock ISSN 1748-6645, 1748-6653.
\newblock \doi{10.1111/jnp.12302}.
\newblock URL
  \url{https://bpspsychub.onlinelibrary.wiley.com/doi/10.1111/jnp.12302}.

\bibitem[Ye and Keogh(2011)]{ye_time_2011}
Lexiang Ye and Eamonn Keogh.
\newblock Time series shapelets: a novel technique that allows accurate,
  interpretable and fast classification.
\newblock \emph{Data Mining and Knowledge Discovery}, 22\penalty0 (1):\penalty0
  149--182, January 2011.
\newblock ISSN 1573-756X.
\newblock \doi{10.1007/s10618-010-0179-5}.
\newblock URL \url{https://doi.org/10.1007/s10618-010-0179-5}.

\end{thebibliography}

\appendix

\section{Data Prepossessing}\label{apd:preprocessing}

The raw data from the eye-tracker measured the participant’s gaze as the angular deviation from the center of the screen for each eye. The angular data was converted to positional data using information about the experimental setup. The left and right eye positions were averaged to create one $x$ component and one $y$ component. The $x$ and $y$ channels were then calibrated by subtracting the mean of the first 300 data points. Data points for which the absolute position was more than three standard deviations away from the mean position over the session were removed and replaced by linearly interpolated values from neighboring data points. We subsequently computed the velocity for the x and y components, obtaining a 4-dimensional time series composed by $x$ position, $y$ position, $x$ velocity and $y$ velocity. 

\paragraph{Data sanitization}
In order to identify and remove corrupted or noisy recordings, the sessions were excluded using a criteria based on the Median Absolute Deviation with respect to the max absolute value and the mean absolute velocity for all sessions. The MAD is a measure of the variability in the data, which is more robust to outliers than standard deviation. A threshold of 3 MADs was used to exclude sessions, and the exclusion was done independently on the x and y channels. The data sanitization removed all data from 21 subjects, resulting in a total of 63 subjects in the dataset.

\paragraph{Splitting Sessions into Trials}
Every MTS corresponding to a session consists of a number of trials. We only used data from the period before the target moves, when the subject is focusing in preparation for the saccade. The time interval length was the same for all trials, and it was such that all of them have a separation of at least $1s$ from the previous target movement, ensuring that the gaze has had time to return to its resting position. Ultimately, the MTS corresponding to trials were centered by substracting the mean to counteract drift throughout the session. A representation of the preprocessing pipeline is shown in figure \ref{fig:preprocessing}. The final dataset consisted of 1584 MTS segments of length 440 ($\sim1.5s$). The MTS segments correspond to trials performed by a subject with one of three conditions: 451 trials from HC, 561 trials from PD patients on their regular dose of dopamine replacement therapy (PD ON), and 572 trials from PD patients after a 12 hour withdrawal from their dopaminergic medication (PD OFF). These trials corresponded to 23 healthy controls, 28 PD patients on medication, and 30 PD patients off medication.

\section{Model Implementation}\label{apd:model}

\subsection{Model 1: InceptionTime}
\label{sec:it}

Our InceptionTime model was implemented in Pytorch Lightning, closely following the original authors’ TensorFlow implementation. The InceptionTime model consisted of 4 Inception modules, each with a bottleneck dimensionality of 32, a hidden dimensionality of 64, and three sets of kernels of sizes [40, 20, 10]. The output of each Inception module is thus $4 \times 64 = 256$ (incl. the parallel max pooling). Following the design of InceptionTime, we use residual connections between the Inception modules to mitigate the vanishing gradient problem. The InceptionTime network had a total of 850k trainable parameters.

\paragraph{Training Setup}
The optimizer used for training InceptionTime was AdamW $(\beta_1=0.9, \beta_2=0.999)$ with a learning rate of \num{1e-3} and a weight decay of \num{1e-3}. The learning rate was scheduled with cosine annealing and 15 epochs of gradual warm up [\cite{goyal_accurate_2018}]. Training was performed for 300 epochs with a large batch size of 256. We saved the model weights after every epoch, and ultimately used the model instance with the best validation set uF1-score. We used the binary cross entropy as loss function. The random state was set before each training run, ensuring the best possible reproducibility. However, the results still varied slightly between runs, due to the stochastic optimization process and GPU accelerated training. The training was performed on a GTX 1080 Ti and took roughly 5 minutes.

\subsection{Model 2: ROCKET}
\label{sec:rocket}

We use an open source implementation of ROCKET from the tsai library\footnote{tsai - A state-of-the-art deep learning library for time series and sequential data (2022) \href{https://github.com/timeseriesAI/tsai}{https://github.com/timeseriesAI/tsai}}. The ROCKET stage uses 10000 kernels to produce as many feature maps. Following computation of the max values and proportion of positive values, the output attributes are normalized to have zero mean and unit standard deviation. The normalized output features were fed to a ridge regression classifier with a ridge parameter of \num{1e4}.

\section{Hyperparameter Results}\label{apd:hyperparam}

Numerical results of hyperparameter search for InceptionTime and ROCKET models are presented in table \ref{tab:it_hyperparam_search} and table \ref{tab:rocket_param_search} respectively.

\begin{table}[h!]
    \centering

        \begin{tabular}{@{}rrrr@{}}
        \multicolumn{1}{l}{}  & \multicolumn{1}{l}{}     & \multicolumn{2}{c}{Learning Rate}                           \\
        \multicolumn{1}{l}{}  & \multicolumn{1}{l|}{}    & \multicolumn{1}{c}{\num{5e-4}} & \multicolumn{1}{c}{\num{1e-3}} \\ \cmidrule(l){2-4} 
        \multirow{4}{*}{\rotatebox[origin=c]{90}{Batch Size}} & \multicolumn{1}{r|}{128} & 0.5738                       & 0.5674                       \\
                              & \multicolumn{1}{r|}{256} & 0.6571                       & \textbf{0.6894}              \\
                              & \multicolumn{1}{r|}{512} & 0.6558                       & 0.6617                                        
        \end{tabular}
    \vspace{1em}
    
        \begin{tabular}{@{}rrrrr@{}}
        \multicolumn{1}{l}{}  & \multicolumn{1}{l}{}   & \multicolumn{3}{c}{Hidden Dimensions}    \\
        \multicolumn{1}{l}{}  & \multicolumn{1}{l|}{}  & 32     & 64              \\ \cmidrule(l){2-5} 
        \multirow{4}{*}{\rotatebox[origin=c]{90}{Depth}} & \multicolumn{1}{r|}{3} & 0.6636 & 0.6550          \\
                              & \multicolumn{1}{r|}{4} & 0.6623 & \textbf{0.6925} \\
                              & \multicolumn{1}{r|}{5} & 0.6786 & 0.6634        
        \end{tabular}
    
    \caption{The results from the hyperparameter search with IncepionTime. The reported numbers are the max uF1-Scores averaged over a 5-fold cross validation.}
    \label{tab:it_hyperparam_search}
\end{table}
\begin{table}[h!]
\centering
\begin{tabular}{@{}rrrrr@{}}
\multicolumn{1}{l}{}  & \multicolumn{1}{l}{}       & \multicolumn{3}{c}{Ridge Parameter} \\
\multicolumn{1}{l}{}  & \multicolumn{1}{l|}{}     & \num{1e-2} & \num{1e-3}   & \num{1e-4}      \\ \cmidrule(l){2-5} 
\multirow{3}{*}{\rotatebox[origin=c]{90}{Kernels}} & \multicolumn{1}{r|}{100}   & 0.5508     & 0.5588     & 0.5617    \\
                      & \multicolumn{1}{r|}{1000}  & 0.5526     & 0.5481     & 0.5369   \\
                      & \multicolumn{1}{r|}{10000} & 0.5897     &\textbf{0.5945}      & 0.5696  
\end{tabular}
\caption{The results from the hyperparameter search with ROCKET. The reported numbers are the max uF1-Scores averaged over a 5-fold cross validation.}
\label{tab:rocket_param_search}
\end{table}
\end{document}